\documentclass[nohyperref]{article}

\usepackage{microtype}
\usepackage{graphicx}
\usepackage{subcaption}
\usepackage{booktabs} 
\usepackage{tabularx}

\usepackage{hyperref}

\usepackage[preprint]{icml2022}

\usepackage{amsmath}
\usepackage{amssymb}
\usepackage{mathtools}
\usepackage{amsthm}
\usepackage{soul}
\usepackage{enumitem}
\usepackage{xspace}

\usepackage[capitalize,noabbrev]{cleveref}

\theoremstyle{plain}

\theoremstyle{definition}

\theoremstyle{remark}

\usepackage{multirow}
\usepackage{multicol}
\usepackage{array}
\usepackage{tikz}
\usepackage{stmaryrd}
\usepackage{xcolor}

\newcommand{\gptj}{GPT-J-6B\xspace}
\newcommand{\neoS}{GPT-Neo-1.3B\xspace}
\newcommand{\neoL}{GPT-Neo-2.7B\xspace}
\newcommand{\metric}{performance gap\xspace}

\newcommand{\freq}{\omega}
\newcommand{\Freq}{\Omega}
\newcommand{\freqx}{\freq_{\{x_1\}}}
\newcommand{\freqy}{\freq_{\{x_1, x_2\}}}
\newcommand{\freqz}{\freq_{\{x_1, y\}}}
\newcommand{\freqxt}{\freq_{\{x_1, x_2\}}}
\newcommand{\freqyt}{\freq_{\{x_1, x_2, x_3\}}}
\newcommand{\freqzt}{\freq_{\{x_1, x_2 ,y\}}}

\newcommand{\gap}{\Delta}
\newcommand{\gapx}{\gap_{1}}
\newcommand{\gapy}{\gap_{1,2}}
\newcommand{\gapz}{\gap_{1,y}}
\newcommand{\gapxt}{\gap_{1,2}}
\newcommand{\gapyt}{\gap_{1,2,3}}
\newcommand{\gapzt}{\gap_{1,2,y}}

\definecolor{colOr}{HTML}{CE822C}
\definecolor{colPr}{HTML}{C561FF}
\definecolor{colGr}{HTML}{03A57A}
\definecolor{colBl}{HTML}{169AC5}
\definecolor{colPi}{HTML}{F26385}

\usepackage{pifont}
\renewcommand{\checkmark}{\ding{51}}
\newcommand{\crossmark}{\ding{55}}

\newif\ifcomments
\commentstrue
\ifcomments
    \providecommand{\yasaman}[2][]{{\protect\color{blue}{[Yasaman:\textbf{#1} #2]}}}
    \providecommand{\rob}[2][]{{\protect\color{red}{[Rob:\textbf{#1} #2]}}}
    \providecommand{\matt}[2][]{{\protect\color{magenta}{[Matt:\textbf{#1} #2]}}}
    \providecommand{\sameer}[2][]{{\protect\color{violet}{[Sameer:\textbf{#1} #2]}}}
    \providecommand{\topic}[2][]{{\protect\color{brown}{[Topic:\textbf{#1} #2]}}}
\else
    \providecommand{\yasaman}[2][]{}
    \providecommand{\rob}[2][]{}
    \providecommand{\matt}[2][]{}
    \providecommand{\sameer}[2][]{}
    \providecommand{\topic}[2][]{}
\fi

\icmltitlerunning{Impact of Pretraining Term Frequencies on Few-Shot Reasoning}

\begin{document}

\twocolumn[
\icmltitle{Impact of Pretraining Term Frequencies on Few-Shot Reasoning}



\icmlsetsymbol{equal}{*}

\begin{icmlauthorlist}
\icmlauthor{Yasaman Razeghi}{uci}
\icmlauthor{Robert L. Logan IV}{uci}
\icmlauthor{Matt Gardner}{comp}
\icmlauthor{Sameer Singh}{uci,ai2}
\end{icmlauthorlist}

\icmlaffiliation{uci}{Department of Computer Science, University of California, Irvine, USA}
\icmlaffiliation{comp}{Microsoft Semantic Machines, USA}
\icmlaffiliation{ai2}{Allen Institute for Artificial Intelligence, USA}

\icmlcorrespondingauthor{Yasaman Razeghi}{yrazeghi@uci.edu}


\vskip 0.3in
]



\printAffiliationsAndNotice{}  

\begin{abstract}



Pretrained Language Models (LMs) have demonstrated ability to perform numerical reasoning by extrapolating from a few examples in few-shot settings.
However, the extent to which this extrapolation relies on robust reasoning is unclear.
In this paper, we investigate how well these models reason with terms that are less frequent in the pretraining data.
In particular, we examine the correlations between the model performance on test instances and the frequency of terms from those instances in the pretraining data.  
We measure the strength of this correlation 
for a number of GPT-based language models (pretrained on the Pile dataset) on various numerical deduction tasks (e.g., arithmetic and unit conversion). 
Our results consistently demonstrate that models are more accurate on instances whose terms are more prevalent, in some cases above $70\%$ (absolute) more accurate on the top 10\% frequent terms in comparison to the bottom 10\%.
Overall, although LMs exhibit strong performance at few-shot numerical reasoning tasks, our results raise the question of how much models actually generalize beyond pretraining data, and we encourage researchers to take the pretraining data into account when interpreting evaluation results.



\end{abstract}

\section{Introduction}
\label{introduction}
\begin{figure}[t]
\centering
\framebox{
\begin{minipage}{0.85\columnwidth}
Q: What is \textcolor{colPr}{$24$} times \textcolor{colPi}{$18$}? A: \makebox[5mm]{\hrulefill}\hspace{5mm} \emph{Model:} \textcolor{colBl}{$432$} \checkmark
\\
Q: What is \textcolor{colPr}{$23$} times \textcolor{colPi}{$18$}? A: \makebox[5mm]{\hrulefill}\hspace{5mm} \emph{Model:} \textcolor{colBl}{$462$} \crossmark
\end{minipage}
}
\includegraphics[width=0.9\columnwidth]{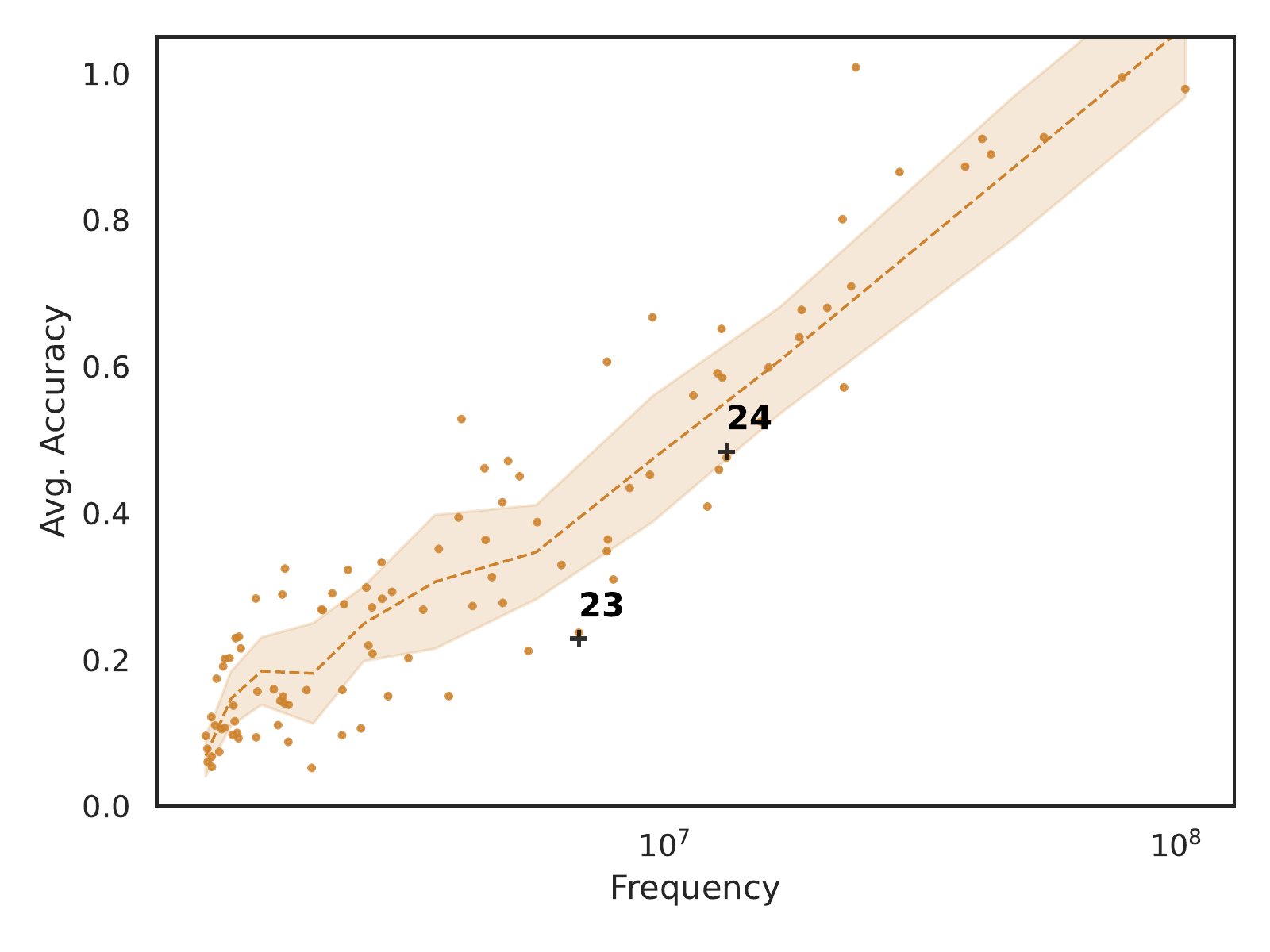}
\vskip -5mm
\caption{
    {\bf Multiplication Performance:} Plot of \gptj's 2-shot accuracy on multiplication (averaged over multiple multiplicands and training instances) against the frequency of the equation's first term in the pretraining corpus.
    Each point represents the average performance for that term (e.g., $24$) multiplied by numbers 1-50 and 5 choices of random seeds. 
    As in the example, the performance difference for the numbers $24$ and $23$ is more than 20\%.
    We find a strong correlation between accuracy and frequency.}
\label{fig:hook}
\end{figure}



Large language models have demonstrated outstanding performance in zero- and few-shot learning settings on numerous tasks, from simple classifications such as sentiment analysis to complex reasoning-related task like natural language inference and arithmetic~\cite{brown2020language, radford2019language}.  
These results suggest that models may have gained the ability to perform simple inductive reasoning through a combination of pretraining and model size.

However, current evaluation schemes for the reasoning of large language models, often neglect or underestimate the impact of data leakage from pretraining data when assessing their reasoning ability. 
Although the overlap between the training and evaluation splits of public datasets and their effect on the generalization of the language models have been studied~\cite{elangovan-etal-2021-memorization, lewis2020question}, the effect of the pretraining data has gotten less attention. 
Traditionally,
a model that has \emph{learned} to reason in the training phase should be able to generalize outside of the narrow context that it was trained in.
Specifically, if the model has learned to reason, its performance on instances with less frequent terms (based on pretraining data) should not be significantly lower than its performance on the instances with more common terms.

As an illustration, consider the arithmetic task of multiplying two integers (shown in Figure~\ref{fig:hook}).
A model that has learned proper arithmetic skills should be able to answer the queries irrespective of the frequencies of the operands in the pretraining data.
Therefore,  it should have roughly equivalent performance when answering the queries \emph{Q: what is $24$ times $X$?} and \emph{Q: what is $23$ times $X$?}, when aggregated over various values of $X$.  This is not the case with current LMs and we will study the effect of frequency terms in details through this paper.
To show the effect of frequency, in this example, we plot the average accuracy of \gptj \cite{mesh-transformer-jax} on the numbers 0--100 (averaged over 1--50 as the other operand) against the frequency of the number in the pretraining data in Figure \ref{fig:hook}.
We find a strong correlation between the term frequency and the model performance indicating that the model reasoning is not robust to these frequencies. 
Note that even ``rare'' terms are overall frequent (on the order of millions) in the pretraining data.  

In this work,  we investigate this impact of the frequency of test instance terms in the model's pretraining data on model's performance. 
We focus our analysis on numerical reasoning tasks, including addition, multiplication, and unit conversion.  
For each of these tasks, we identify relevant \emph{terms} from each instance; for these tasks, terms are the numbers and units involved.
We count occurrences of these terms in the pretraining data, including co-occurrences of term pairs or triples within a fixed window.  
This procedure allows us to aggregate over instances in which these terms appear and observe the relationship between term frequency and model accuracy on instances that include those terms. 
We summarize this behavior through the \emph{performance gap} between instances that have the most frequent terms and instances that have the least frequent terms.
Intuitively, models that exhibit a \textit{high} performance gap are \textit{more} accurate on instances that are more common in the pretraining data; this indicates that the model does not generalize appropriately and is likely affected by dataset overlap.

We present analysis on these numerical reasoning tasks for three sizes of the EleutherAI/GPT models pretrained on the Pile~\cite{pile} dataset, which has been publicly released and thus permits this kind of analysis (in contrast to the data that, e.g., GPT-3 \cite{brown2020language} was trained on).
Our results show a consistently large performance gap between highest-frequency terms and lowest-frequency terms in all of our experiments; in some cases there is more than $70\%$ of average accuracy gap between the first and last $10\%$ indicating that even simple unigram statistics are highly correlated with models performance which should not happen if the model is performing reasoning. 
These observations suggest that any evaluation of reasoning that does not take the pretraining data into account is difficult to interpret, and that we need to revisit evaluation of language models with respect to their pretraining data.
\section{Background and Methodology}

Reasoning ability has long been considered as a proxy for intelligence~\cite{Johnson-Laird18243}. 
Thus, developing models with this skill has been also an essential goal of AI and natural language processing (NLP)~\cite{bommasani2021opportunities}.
Recently, large language models have exhibited an ability to perform reasoning-related tasks in few-shot settings without requiring any modifications to their parameters through a method called in-context learning.
Our goal is to evaluate this reasoning skill in-depth for numerical induction tasks.
This section provides background information on in-context learning and introduces our method for measuring the performance gap of the models on numerical reasoning tasks based on differences in pretraining term frequency.


\subsection{In-context Learning}
\citet{brown2020language} show that the large GPT-3 model is able to perform well on few-shot reasoning tasks without requiring any changes to its internal parameters, through the usage of a technique called \emph{in-context learning}.
In place of a typical learning procedure, in-context learning instead places training examples in a prompt format, which is subsequently fed to a language model as its input. 

Among numerous experiments, \citet{brown2020language} show that GPT3 performs well on a variety of arithmetic questions such as addition and subtraction with 2--5 digit numbers.
For example, they show that their largest model can perform zero-shot 2-digit addition with $76.9\%$ accuracy. 
Although impressive, due to the large volume of data GPT-3 is trained on, it is possible that the model is merely repeating answers seen during pretraining.
To attribute this performance to the model’s  \emph{reasoning} capabilities, we need to make sure that the model is not affected by statistical overlaps between the terms of the arithmetic questions and the pretraining data.

In the following sections, we introduce metrics that we use to investigate the relationship between the frequency of terms in the pretraining data and the model performance on reasoning instances containing those terms.
To assess this relation, we first define an approach for measuring term frequencies in a large pretraining dataset (Section~\ref{subsection:frequency}). 
We connect these frequencies to reasoning performance by introducing the \emph{performance gap} $\gap$ (Section~\ref{subsection:gap}).


\subsection{Frequency}
\label{subsection:frequency}



We consider numerical reasoning tasks (Table~\ref{tab:example-templates}) whose instances consist of input terms, $\mathbf{x}=(x_1,\ldots, x_i, \ldots x_{n})$, and a derived output term $y$, where the $x_i$'s are either positive integers or units of time (e.g., 1, 2, {\tt hour}, etc.) and $y$ is a positive integer. 
For example, for the task of multiplication, an instance might be $\mathbf{x}=(23, 18)$ and $y=414$, representing the equation $23\times  18=414$.

For each instance, we extract counts of the number of times that a subset of its terms $X\subseteq \left\{ x_1, \ldots, x_n, y \right\} $ appear within a specified window in the pretraining data.
We refer to this count as the frequency, ${\freq}_{X}$, of $X$.

In this paper, we restrict our attention to frequencies involving three or less input terms, e.g., $\mathbf{x} = (x_1)$ or $(x_1, x_2)$ or $(x_1, x_2, x_3)$ and optionally the output term $y$, e.g.:
\begin{itemize}[nosep,leftmargin=*]
     \item $\freqx$: the number of times that $x_1$ (e.g., $23$) appears in the pretraining data.
     \item $\freqy$: the number of times that the input terms $x_1$ (e.g., $23$) and $x_2$ (e.g., $18$) appear in the pretraining data within a specific window size.
     \item $\freqz$: the number of times that the first input term $x_1$ (e.g., $23$) and the output term $y$ (e.g., $414$) appear in the pretraining data within a specific window size.
\end{itemize}
Note that our usage of set notation in the subscript is deliberate; although $\mathbf{x} = (x_1, x_2)$ and $\mathbf{x}' = (x_2, x_1)$ are not necessarily the same (e.g., order is important when representing the task instance), frequency is symmetric (e.g., $\freq_{\{x_1, x_2\}} = \freq_{\{x_2, x_1\}} \forall x_1, x_2$).



\subsection{Performance Gap}
\label{subsection:gap}

We want to measure how much \emph{more} accurate the model is on instances containing more versus less frequent terms in the pretraining data.
We do this by calculating the differences in average accuracies of the instances in the top and bottom quantiles of the distribution over term frequencies, which we call the \emph{performance gap}.

Formally, let $\{(X^{(n)}, \freq_X^{(n)})\}$, $n \in [1, N]$, be a set of terms for a task and their associated term frequencies in the pretraining corpus.
Given a task (e.g. addition), we create reasoning instances for each element of this set by instantiating uninstantiated values of $x_i$, and deriving $y$ if $y \notin X^{(n)}$.
We then measure the LM's accuracy $a^{(n)}$ over the set of instances, and repeat this process for all $n \in [1,N]$, producing a set $\Freq = \{(\freq_X^{(n)}, a^{(n)})\}$.
The formula for the {\bf performance gap} is then given by:
\begin{equation}
    \gap(\Freq) = \text{Acc}(\Freq_{>90\%}) - \text{Acc}(\Freq_{<10\%})
    \label{eq:performance_gap}
\end{equation}
where $\Freq_{>90\%}$ is the top 10\% of elements in $\Freq$ ordered by frequency, $\Freq_{<10\%}$ is the bottom 10\%, and $\text{Acc}(\Freq')$ is the average accuracy of elements in $\Freq'$.
We introduce the following convenient abuses of notation $\gap_1$, $\gap_{1,2}$, $\gap_{1,y}$, \ldots, to denote the performance gap over the frequency distributions of $\freqx$, $\freqy$, $\freqz$, \ldots, respectively.

Concretely, for the multiplication example from Figure~\ref{fig:hook}, $\mathbf{x}=(x_1,x_2)$ and we consider the performance gap over frequencies $\freqx$.
For each number (say $23$), we count the number of times it appears in the pretraining corpus ($\freq_{\{23\}}$), and compute the average accuracy of the model over all instances where the first operand is $23$.
The performance gap w.r.t. to $\freqx$ for this task is the difference between the average accuracy over the top 10\% and the bottom 10\% most frequent numbers in the pretraining corpus.

\begin{table}[t]
    \caption{Prompt templates and the number of test cases investigated for each numerical reasoning task.}
    \label{tab:example-templates}
    \vskip 0.15in
    \small
    \centering
    \def\tabcolsep{3.5pt}
    \begin{tabularx}{\linewidth}{l l r} 
        \toprule
        \multirow{2}{*}{\bf Task} & \multirow{2}{*}{\bf Prompt Template}   & \bf\#Test \\
                              &  & \bf Cases \\
        \midrule
        \multicolumn{3}{l}{\bf Arithematic} \\
        Multiplication       & \em Q:What is \textcolor{colPr}{$x_1$} times \textcolor{colPi}{$x_2$}?  A: \textcolor{colBl}{$y$} & $5000$\\
        Addition        & \em Q:What is \textcolor{colPr}{$x_1$} plus \textcolor{colPi}{$x_2$}?  A: \textcolor{colBl}{$y$} & $5000$\\
                    \addlinespace
        \multicolumn{3}{l}{\bf Operation Inference} \\
        Mult. \#    & \em Q:What is \textcolor{colPr}{$x_1$} \# \textcolor{colPi}{$x_2$}?  A: \textcolor{colBl}{$y$}& $5000$ \\
        Add. \#     & \em Q:What is \textcolor{colPr}{$x_1$} \# \textcolor{colPi}{$x_2$}? A: \textcolor{colBl}{$y$}& $5000$ \\
                    \addlinespace
        \multicolumn{3}{l}{\bf Time Unit Inference} \\
        Min$\shortrightarrow$Sec     & \em Q:What is \textcolor{colPr}{$x_1$} \textcolor{colPi}{minutes} in seconds? A: \textcolor{colBl}{$y$}& $79$\\
        Hour$\shortrightarrow$Min    & \em Q:What is \textcolor{colPr}{$x_1$} \textcolor{colPi}{hours} in minutes? A: \textcolor{colBl}{$y$}& $100$ \\
        Day$\shortrightarrow$Hour    & \em Q:What is \textcolor{colPr}{$x_1$} \textcolor{colPi}{days} in hours? A: \textcolor{colBl}{$y$}& $100$ \\
        Week$\shortrightarrow$Day    & \em Q:What is \textcolor{colPr}{$x_1$} \textcolor{colPi}{weeks} in days? A: \textcolor{colBl}{$y$}& $100$ \\
        Month$\shortrightarrow$Week  & \em Q:What is \textcolor{colPr}{$x_1$} \textcolor{colPi}{months} in weeks? A: \textcolor{colBl}{$y$}& $100$ \\
        Year$\shortrightarrow$Month  & \em Q:What is \textcolor{colPr}{$x_1$} \textcolor{colPi}{years} in months? A: \textcolor{colBl}{$y$}& $100$\\
        Decade$\shortrightarrow$Year & \em Q:What is \textcolor{colPr}{$x_1$} \textcolor{colPi}{decades} in years? A: \textcolor{colBl}{$y$}& $100$\\
        \bottomrule
    \end{tabularx}
    \vskip -0.1in
\end{table}

\begin{figure*}[tb]
    \centering
    \includegraphics[width=\textwidth]{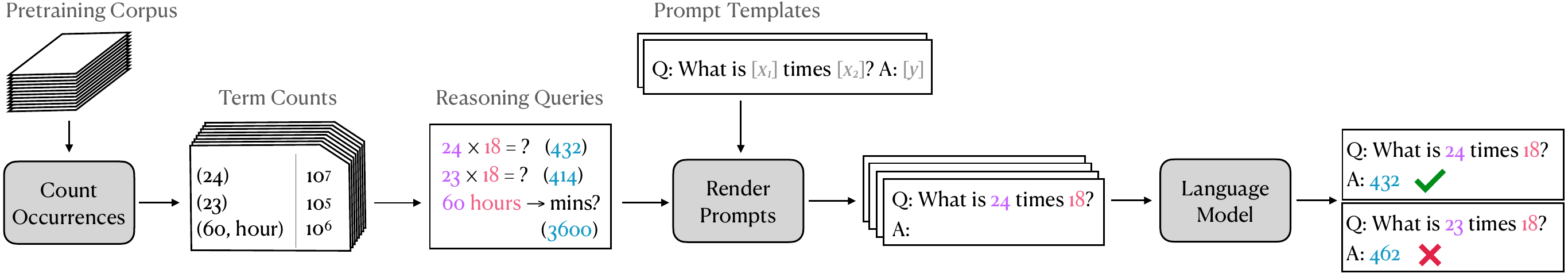}
    \caption{\textbf{Pipeline for Data Construction:} We use the term counts processed from the pretraining data to develop the reasoning queries and render them with prompts templates to a proper language model input format (illustrated using the example from Figure~\ref{fig:hook}).}
    \label{fig:pipeline}
\end{figure*}

\section{Experiment Setup}

In this section, we describe our setup to measure the effect of pretraining data on the few-shot evaluation of a number of numerical reasoning tasks for different language models.
For reproducibility, we will release the complete source code of our experiments.

\paragraph{Language Models}
We experiment on models from EleutherAI i.e., \gptj~\cite{mesh-transformer-jax}, and \neoS, \neoL~\cite{gpt-neo}.
These models are publicly available, but more importantly, the corpus used to pretrain them has also been released. 
We use the HuggingFace\footnote{Source code at {\scriptsize \url{https://huggingface.co/EleutherAI}}} Transformer integration  of the models in our experiments.

\paragraph{Pretraining Corpus}
Language models studied in this work are trained on the Pile dataset~\cite{pile}, a large-scale language modeling dataset (800GB) consisting of English documents in 22 academic or other professional data sources. 
Since our reasoning tasks focus on numbers, we count the frequency of all integers with less than seven digits using a white-space tokenizer.
To calculate the frequencies of the numbers we use Amazon Elastic Map Reduce (EMR) platform.
We will release the code and statistics of the frequencies.

\paragraph{Numerical Reasoning Tasks}
We create three types of datasets that target the mathematical capabilities of language models
since solving mathematical questions is a useful \emph{reasoning} capability of the models~\cite{brown2020language}.
\begin{itemize}[itemsep=1mm,topsep=0mm,leftmargin=*]
    \item {\bf Arithmetic, 2 tasks} 
    As the first task, we consider simple arithmetic operations: addition $x_1+x_2\shortrightarrow y$ and multiplication $x_1\times x_2\shortrightarrow y$.
    In both cases, the first operands ($x_1$) are numbers less than $100$ (that are in the top $200$ most frequent numbers) and the second operands ($x_2$) are the numbers in the range ($1-50$). 

    \item {\bf Operation Inference, 2 tasks} 
    Instead of directly specifying the operation, we also create a variation where the model needs to \emph{infer}, from a few examples, the operation itself, as well as the result, as introduced in the evaluation of Megatron-Turing model~\footnote{\scriptsize \url{https://turing.microsoft.com/}}. 
    We replace the arithmetic operation with a ``\#'', with the same operations and operands as previous, to create these datasets. 

    \item {\bf Time Unit Conversion, 7 tasks}
    Apart from directly executing arithmetic expressions, we are also interested in evaluating model capability to implicitly reason about these operations.
    To this end, we construct a unit conversion dataset by identifying the most frequent numbers that co-occur with time unit words (``second'', ``minute'', ``hour'', ``day'', ``week'', ``month'', ``year'', and ``decade'') as the primary operand $x_1$, the time units themselves as additional operands ($x_2\shortrightarrow x_3$), i.e. converting $24$ hours to minutes is represented as ($24$, ``hours'', $60$). 
    We expect converting time values to be mathematically more straightforward than two-digit multiplication since the model need only multiply with the same (implicit) second operand, e.g., $\times 60$ for converting hours to minutes. 
\end{itemize}
The overall pipeline for creating natural language instances for these datasets is illustrated in Figure~\ref{fig:pipeline}.
We compute occurrences and co-occurrences (if they are within a window of $5$ words) of the terms in the corpus, i.e. the time units and numbers.
We then generate instances for each of our reasoning datasets using the most frequent terms (in the top $200$) as operands with less than 3 digits.
We focus on the top terms since we expect the models to have a fairly reliable and robust representations for these words.
Each reasoning instance is \emph{rendered} as a natural language query using the prompt templates from Table~\ref{tab:example-templates}, and fed to the language model to generate the answer.
For example, to create a multiplication instance given the argument $(x_1=23, x_2=18)$, we use the instance template to create a natural language input for the model as \emph{``Q: What is $23$ times $18$? A: \makebox[3mm]{\hrulefill}''}, with the goal of producing ``$414$'' ($y=23\times 18=414$).
For few-shot evaluation, we prompt the language models with $k=0, 2, 4, 8, 16$ shots, and average performance over five random selection of the prompt instances, we randomly select $k$ samples from the dataset as the training examples provided in the prompt and  use the rest of the instances from the model as the test data.



\section{Results}

With the three types of reasoning tasks (consisting of 11 total datasets), we present an evaluation of the effect of pretraining frequency on the performance of the language models.
For each dataset, we measure the performance gap on instances that consist of rarer (relatively) terms, for a few different choices of what to compute frequency over i.e different combinations of the instance terms.
To observe this effect with larger language models, we also investigate the effect of the model size on this performance gap.


\begin{figure*}[t]
    \centering

    \begin{subfigure}[b]{0.24\textwidth}
        \centering
        \includegraphics[width=\textwidth,clip,trim=0 0 425 0]{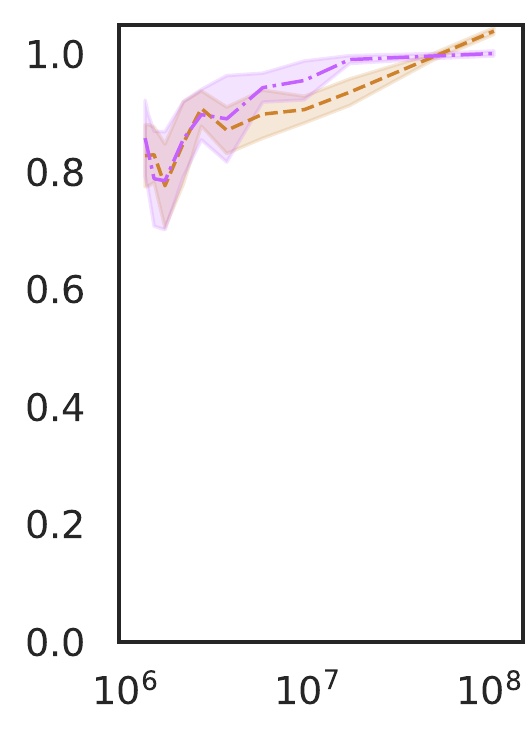}
        \caption{Arithmetic-Addition}
        \label{fig:results:gptjmodel:add}
    \end{subfigure}
    \begin{subfigure}[b]{0.24\textwidth}
        \centering
        \includegraphics[width=0.935\textwidth,clip,trim=157 0 278 0]{figures/quadplot.pdf}
        \caption{Arithmetic-Multiplication}
        \label{fig:results:gptjmodel:mult}
    \end{subfigure}
    \begin{subfigure}[b]{0.24\textwidth}
        \centering
        \includegraphics[width=0.935\textwidth,clip,trim=295 0 140 0]{figures/quadplot.pdf}
        \caption{Op.Inference-Addition}
        \label{fig:results:gptj:plushashtag}
    \end{subfigure}
     \begin{subfigure}[b]{0.24\textwidth}
        \centering
        \includegraphics[width=0.935\textwidth,clip,trim=435 0 0 0]{figures/quadplot.pdf}
        \caption{Op. Inference-Multiplication}
        \label{fig:results:gptj:multhashtag}
    \end{subfigure}

    \caption{{\bf The {\gptj} accuracy on arithematic and operator inference} tasks, with $k$ shots. The average accuracy ($y$-axis) of the binned instances is highly correlated with their term frequencies $\freqx$ in the pretraining corpus ($x$-axis).}
    \label{fig:results:gptj:hashtag}
\end{figure*}

\begin{table*}[t]
    \caption{
        {\bf {\gptj} results on arithmetic, operation inference (\#) tasks} $\gapx$, $\gapy$ and $\gapz$ represent the performance gap over the frequency distributions of $\freqx$, $\freqy$ and $\freqz$ respectively. $x_1$ represent the first operand, $x_2$ second operand and $y$ the answer of the arithmetic question.
    }
    \label{tab:results:arith_and_hashtag}
    \vskip 0.15in
    \centering
    \small
    \begin{tabular}{l rrrr rrrr rrrr rrrr}
        \toprule
        \multirow{2}{*}{$k$}
        & \multicolumn{4}{c}{Multiplication}
        & \multicolumn{4}{c}{Addition}
        & \multicolumn{4}{c}{Multiplication (\#)}
        & \multicolumn{4}{c}{Addition (\#)}
        \\
        \cmidrule(lr){2-5}
        \cmidrule(lr){6-9}
        \cmidrule(lr){10-13}
        \cmidrule(lr){14-17}
        & Acc. & {\scriptsize$\gapx$} & {\scriptsize$\gapy$} & {\scriptsize$\gapz$}
        & Acc. & {\scriptsize$\gapx$} & {\scriptsize$\gapy$} & {\scriptsize$\gapz$} 
        & Acc. & {\scriptsize$\gapx$} & {\scriptsize$\gapy$} & {\scriptsize$\gapz$}
        & Acc. & {\scriptsize$\gapx$} & {\scriptsize$\gapy$} & {\scriptsize$\gapz$}
        \\
        \midrule
        0 & 5.4 & 18.0 & 20.6 & 30.8
          & 1.6 & 8.4 & 6.9 & 8.0
          & - & - & - & - 
          & - & - & - & - \\
        2 & 35.9 & 77.6 & 79.3 & 89.9
          & 88.2 & 16.8 & 21.7 & 21.9
          & 3.1 & 14.1 & 13.7 & 14.2
          & 7.8 & 18.1 & 25.3 & 28.3\\
        4 & 39.2 & 70.8 & 76.4 & 83.5
          & 91.4 & 15.0 & 24.8 & 26.4
          & 5.7 & 20.9 & 21.3 & 23.4
          & 9.8 & 24.8 & 30.1 & 30.4\\
        8 & 42.9 & 74.6 & 80.8 & 86.0
          & 89.6 & 16.3 & 26.5 & 29.6
          & 9.4 & 31.3 & 33.2 & 34.7
          & 19.8 & 31.0 & 44.8 & 45.2\\
        16 & 40.9 & 73.3 & 77.7 & 82.6
           & 88.6 & 16.4 & 27.3 & 31.0
           & 11.0 & 39.6 & 38.7 & 42.6
           & 26.2 & 38.5 & 47.2 & 49.9\\
        \bottomrule
    \end{tabular}
    \vskip -0.1in
\end{table*}

\paragraph{Arithmetic}
We first study the performance on simple addition and multiplication of numbers.
The results for the {\gptj} model is provided in Table~\ref{tab:results:arith_and_hashtag}, with performance gap computed just for $x_1$, for $(x_1,x_2)$, and for $(x_1,y)$.
In \emph{multiplication}, we observe a very high {\metric} for all these definitions of frequencies, suggesting a strong effect of frequency in the pretraining data on the model's ability to perform multiplication.
For better illustration of the \metric, we plot the mean accuracy across the frequency of $x_1$ in Figure~\ref{fig:results:gptjmodel:mult}. 
The plot demonstrates the strong correlation between the models accuracy on specific instances, and the instance element frequency in the pretraining data. 
For \emph{addition}, we observe an overall higher performance of the {\gptj} model in comparison to the multiplication experiments. 
However, the {\metric} on all of the definitions of the instance frequencies still shows an strong effect on the models accuracy. 
As shown in Figure~\ref{fig:results:gptjmodel:add}, the average accuracy of the model still has a positive slope, indicating the effect of instance frequencies.


\paragraph{Operation Inference}
These tasks aim to assess the model capability to both infer the math operation and to perform the actual computation. 
Overall, as we see in Table~\ref{tab:results:arith_and_hashtag}, the model is much less accurate here as compared to the arithmetic experiments. 
However, the model has better performance on the frequent instances even for these low performance tasks (see detailed trend in Figures \ref{fig:results:gptj:multhashtag} and \ref{fig:results:gptj:plushashtag}). 
The performance gap here suggests that the effect of pretraining is not only for tasks that the model is accurate on, but even for operation inference that is more challenging and require deeper reasoning.
An additional interesting trend is in the last column of Table~\ref{tab:results:arith_and_hashtag} with the steady growth of {\metric} with respect to $\freqz$ (co-occurrence of the instance input element and the true answer) as the number of shots increases and the model gets better in performance.
Moreover, the lower accuracy here as compared to addition experiments in the previous section suggests that the model is unable to infer the operation from the few-shot prompts, and it may be performing some form of pattern matching based on the pretraining data.

\begin{table*}[t]
    \caption{\textbf{{\gptj} results on Time-Unit Conversion:}
    $\gapxt$, $\gapyt$ and $\gapzt$ represent the performance gap over the frequency distributions of $\freqxt$, $\freqyt$ and $\freqzt$ respectively, where $x_1$ is the number operand, $x_2$ is the source time unit, $x_3$ is the second implicit number operand needed for performing the conversion and the $y$ is the true answer. }
    \label{tab:results:timeunits:alt}
    \vskip 0.15in
    \centering
    \small
    \def\tabcolsep{5pt}
    \begin{tabular}{l rrrr rrrr rrrr rrrr}
        \toprule
        \multirow{2}{*}{$k$}
        & \multicolumn{4}{c}{Min$\shortrightarrow$Sec}
        & \multicolumn{4}{c}{Hour$\shortrightarrow$Min}
        & \multicolumn{4}{c}{Day$\shortrightarrow$Hour}
        & \multicolumn{4}{c}{Week$\shortrightarrow$Day}
        \\
        \cmidrule(lr){2-5}
        \cmidrule(lr){6-9}
        \cmidrule(lr){10-13}
        \cmidrule(lr){14-17}
        & Acc. & {\scriptsize$\gapxt$} & {\scriptsize$\gapyt$} & {\scriptsize$\gapzt$}
        & Acc. & {\scriptsize$\gapxt$} & {\scriptsize$\gapyt$} & {\scriptsize$\gapzt$} 
        & Acc. & {\scriptsize$\gapxt$} & {\scriptsize$\gapyt$} & {\scriptsize$\gapzt$}
        & Acc. & {\scriptsize$\gapxt$} & {\scriptsize$\gapyt$} & {\scriptsize$\gapzt$}
        \\
        \midrule
        0  &  1.3 &  0.0 &  0.0 & 12.5 
           &  1.0 &  0.0 &  0.0 &  5.0
           &  1.0 &  0.0 &  0.0 & 10.0
           &  1.0 &  0.0 &  0.0 & 10.0
            \\
        2  & 25.5 & 62.5 & 67.5 & 67.5 
           & 19.4 & 58.0 & 40.5 & 44.0
           & 12.1 & 28.9 & 24.0 & 28.0
           & 13.1 & 43.5 & 50.0 & 54.0
            \\
        4  & 35.5 & 60.0 & 71.7 & 63.1 
           & 29.1 & 76.4 & 50.5 & 59.0
           & 22.7 & 46.4 & 45.0 & 47.5
           & 19.2 & 40.9 & 43.3 & 47.0
            \\
        8  & 49.9 & 72.1 & 79.0 & 52.7 
           & 36.3 & 74.6 & 52.5 & 63.0
           & 31.0 & 59.1 & 52.5 & 54.5
           & 28.6 & 70.6 & 62.0 & 67.0
            \\
        16 & 58.4 & 82.7 & 74.4 & 48.5 
           & 42.8 & 80.1 & 49.0 & 62.5
           & 43.3 & 62.8 & 56.0 & 54.8
           & 28.0 & 22.1 & 31.4 & 33.2
            \\
         \bottomrule
    \end{tabular}
    \vskip 2mm
   \begin{tabular}{l rrrr rrrr rrrr rrrr}
        \toprule
        \multirow{2}{*}{Shots, $k$}
        & \multicolumn{4}{c}{Month$\shortrightarrow$Week}
        & \multicolumn{4}{c}{Year$\shortrightarrow$Month}
        & \multicolumn{4}{c}{Decade$\shortrightarrow$Year}
        \\
        \cmidrule(lr){2-5}
        \cmidrule(lr){6-9}
        \cmidrule(lr){10-13}
        & Acc. & {\scriptsize$\gapxt$} & {\scriptsize$\gapyt$} & {\scriptsize$\gapzt$}
        & Acc. & {\scriptsize$\gapxt$} & {\scriptsize$\gapyt$} & {\scriptsize$\gapzt$} 
        & Acc. & {\scriptsize$\gapxt$} & {\scriptsize$\gapyt$} & {\scriptsize$\gapzt$}
        \\
        \midrule

        0  &  1.0 &  0.0 &  0.0 & 10.0
           &  1.0 &  0.0 &  0.0 & 10.0
           &   3.1 & 14.3 & 14.3 & 28.6
        \\
        2  & 30.1 &  8.5 &  9.3 & 21.0
           & 21.8 & 58.0 & 64.0 & 53.0
           &  76.5 & 38.8 & 47.1 & 43.1
        \\
        4  & 63.3 & 22.9 & 26.2 & 10.5
           & 31.9 & 64.8 & 69.5 & 66.8
           &  96.7 & 2.9 & 0.0 & 2.9
        \\
        8  & 80.9 & 33.8 & 30.8 & 24.0
           & 45.4 & 55.0 & 72.0 & 50.0
           &  99.6 & 0.0 & 0.0 & 0.0
        \\
        16 & 84.5 & 43.4 & 57.0 & 30.3
           & 56.7 & 58.7 & 65.3 & 61.3
           & 100.0 & 0.0 & 0.0 & 0.0
        \\
        \bottomrule
    \end{tabular}
    \vskip -0.1in
\end{table*}

\begin{figure}[t]
\centering
\includegraphics[width=0.85\columnwidth]{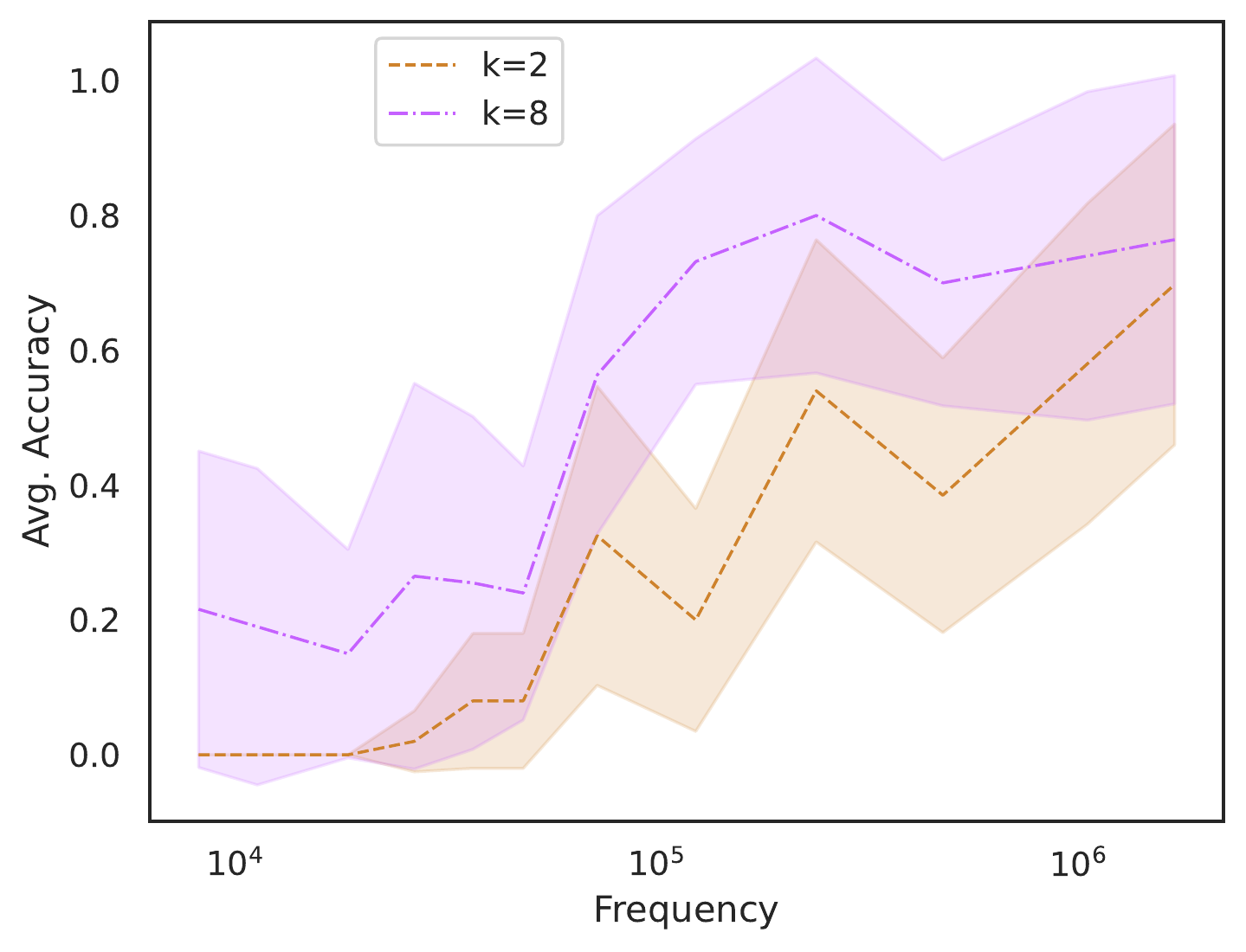}
\caption{ \textbf{{\gptj} performance on Year$\shortrightarrow$Month:} 
 The interpolation lines show the correlation between the average accuracy and the $\freqxt$. $k$ is the number of shots.}
\label{fig:results:gptj:year}
\end{figure}

\begin{figure}[t]
\centering
\includegraphics[width=0.85\columnwidth]{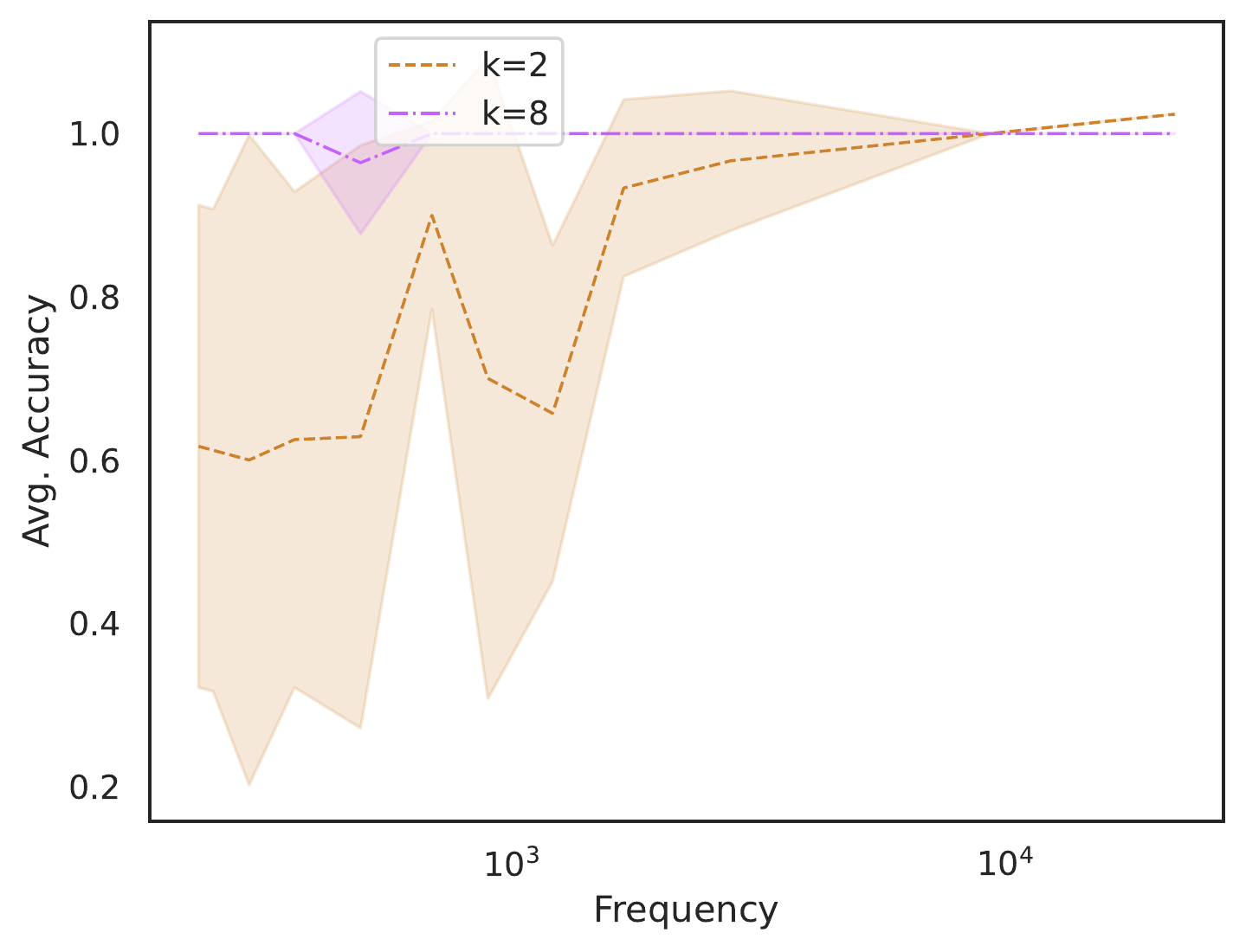}
\caption{ \textbf{{\gptj} performance on Decade$\shortrightarrow$Year:} 
  The interpolation average accuracy line over the $\freqxt$ show that the model reaches a high performance with the number of shots $k=8$, there is still a performance gap in the case of $k=2$. }
\label{fig:results:gptj:decade}
\end{figure}

\paragraph{Time-Unit Conversion}
The {\metric} evaluated on all the time unit conversion experiments is provided in Table~\ref{tab:results:timeunits:alt}. 
We first observe a relatively high {\metric} on all the tasks except the conversion from decade to year. 
For example, Figure~\ref{fig:results:gptj:year} illustrates the trend of increased model performance with the frequency of instance elements for converting the time values from years to months. 
We also observe a general pattern of increase in the {\metric} as the number of shots (training examples in the prompt) increases (results are in table~\ref{tab:results:timeunits:alt}).
These results suggest that even though the model gets more accurate, the improvements focus on more frequent instances of the task.
\emph{Decades to years:}
As we can observe in Figure~\ref{fig:results:gptj:decade}, 
The model performs nearly perfectly on this task with as few as $8$ shots, and we only see very small \metric s.
This is likely due to the task being quite simple (appending a ``0'' to the input number) and the model is able to \emph{generalize} in the manner we are evaluating it.
However, it is also possible that we are simply not identifying the right frequency statistics for this task, and there is an effect that our current evaluation setup does not capture.





\begin{figure}[t]
\centering
    \begin{subfigure}{\columnwidth}
        \centering
        \includegraphics[width=0.85\columnwidth]{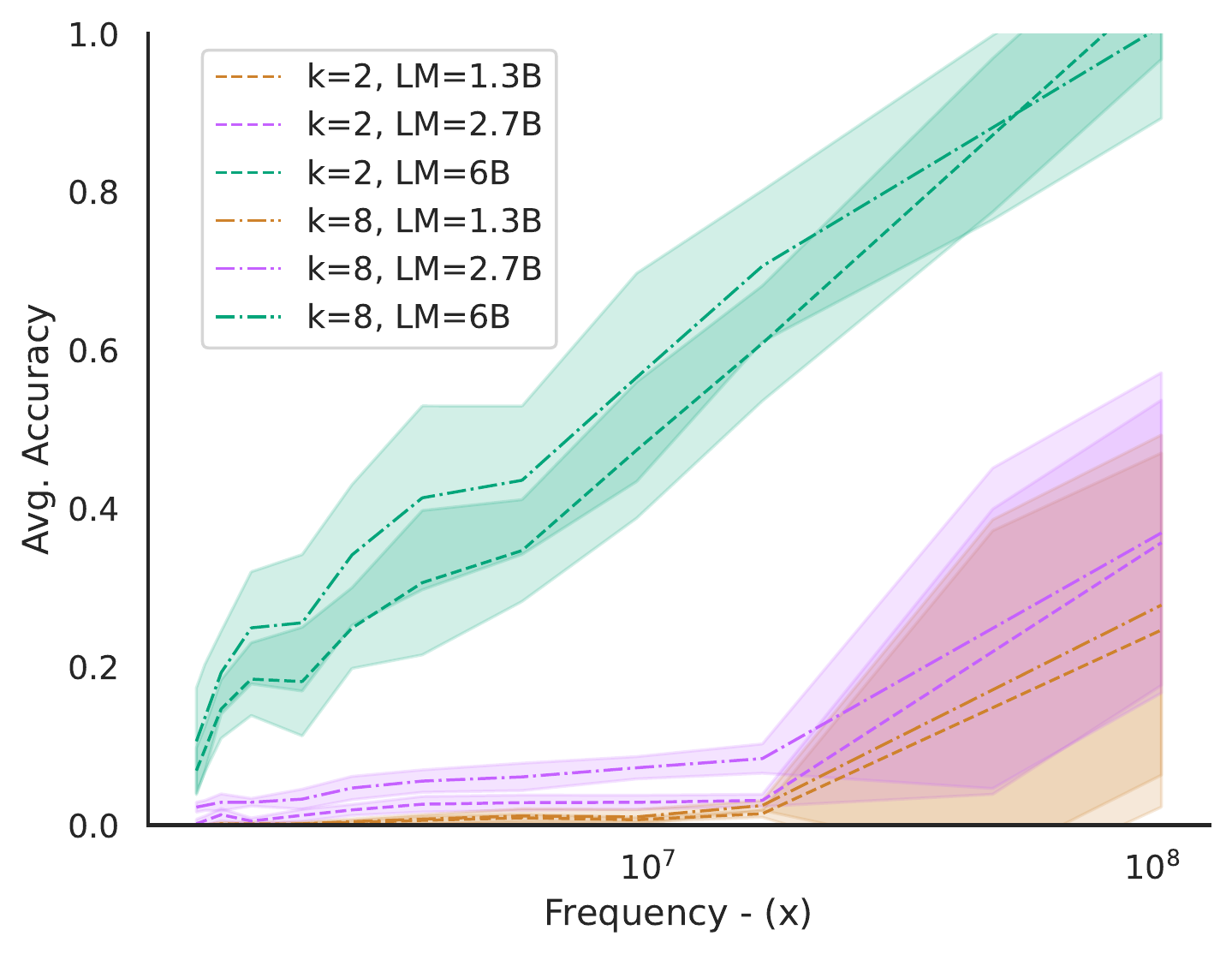}
        \caption{Arithmetic-Multiplication}
        \label{fig:results:modelsizes:mult}
    \end{subfigure}

    \begin{subfigure}{\columnwidth}
        \centering
        \includegraphics[width=0.85\columnwidth]{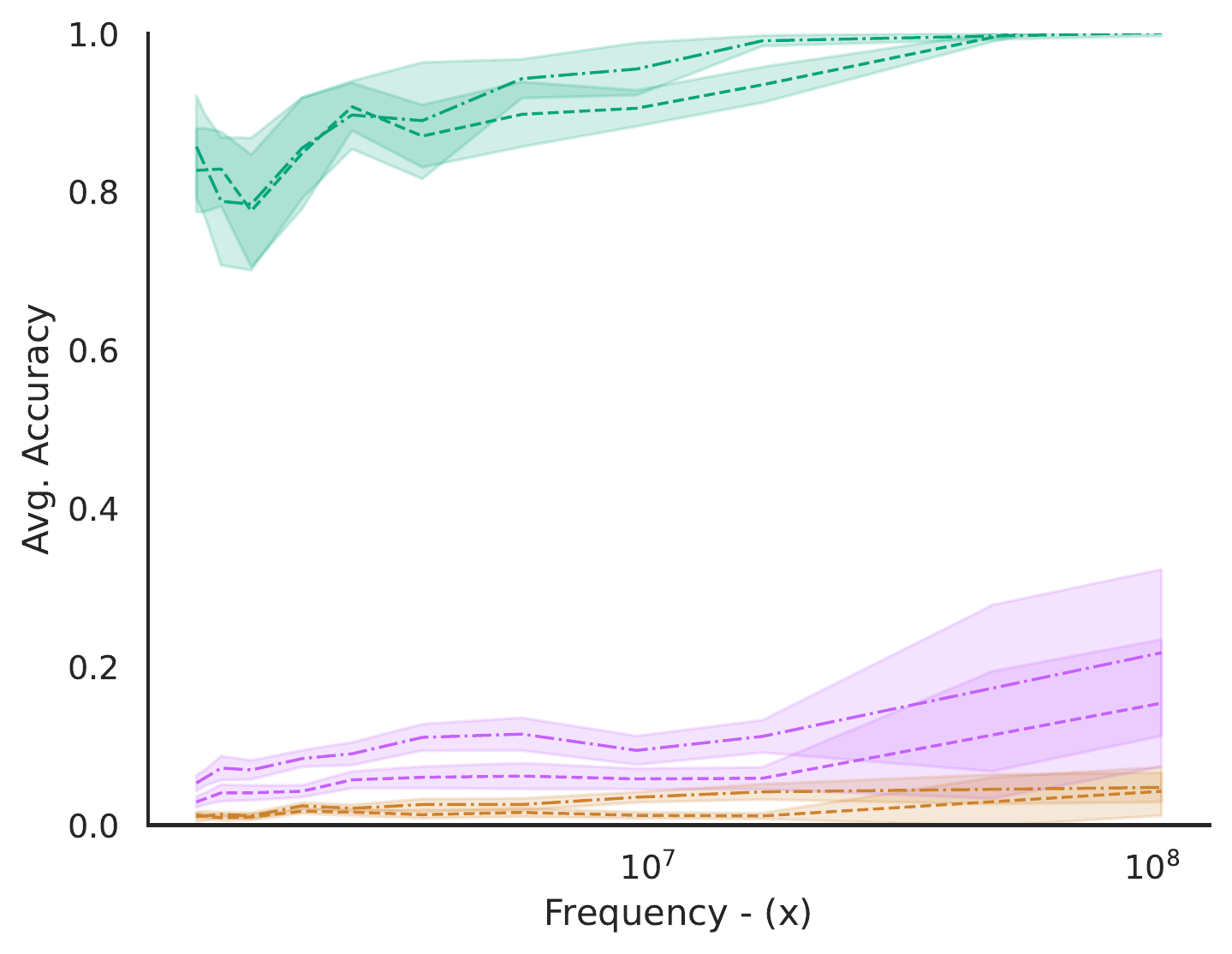}
        \caption{Arithmetic-Addition}
         \label{fig:results:modelsizes:plus}
    \end{subfigure}

\caption{{\bf The effect of model size on performance:} Smaller models only perform well on instances with more frequent terms in the pretraining data. $k$ represents the number of shots.}
\label{fig:results:gptjmodel:modelsizes}
\end{figure}

\paragraph{Studying the Size of Language Models}
To further study the impact of language models sizes on the \metric caused by the instance frequencies, we perform the arithmetic experiments for $2$ and $8$ shots using a variety of models (including the smaller versions of models {\neoS} and {\neoL}). 
We can see the trends of the average accuracy of the models in Figures \ref{fig:results:modelsizes:mult} and \ref{fig:results:modelsizes:plus}. 
The smaller models overall are less accurate on the arithmetic tasks, which is consistent with observations in related work~\cite{brown2020language}. 
However, their success is still focused on the more frequent terms from the pretraining corpus, suggesting that even the smaller models show the effect of reliance on the pretraining data, although to a much lower extent than the larger ones.



\paragraph{Summary}
Overall, we observe high positive {\metric} for almost all of the experiments on the three definition levels of the frequency for each task.
This suggests a strong effect of frequency of the instances in the pretraining data on the model performance.
In particular, evaluation using {\metric} with $\freqx$ shows that even the \emph{unigram} statistics of the instances have strong correlation with the models performance on the instance. 

Other than some exceptional cases, we observe an increasing trend in the performance gap as we put more training instances in the prompt
(the number of shots);
this can be a further indication that the model is directed through the patterns in the pretraining data to answer the reasoning questions. 
Our experiments with the smaller sizes of the model also show that they can only solve the frequent instances of the tasks, which further supports our observation that model performance is correlated with the term frequencies.

\section{Related Work}

A large and growing body of literature has investigated a number of related concerns with large language models.

\paragraph{Prompting} Prompting has been widely applied to study the factual~\cite{petroni2019lama}, commonsense~\cite{davison2019commonsense, weir2020probing, lin2020birds}, mathematical~\cite{saxton2019analysing}, and other NLP task-related~\cite{radford2019language,shin2020autoprompt} knowledge LMs acquire during pretraining.
In this work, we focus on the \emph{in-context learning} setup of \citet{brown2020language}, who use prompts that include training examples to diagnose LMs' few-shot learning capabilities. 


\paragraph{Impact of Frequency on LM Performance}
\citet{kassner2020pretrained} and \citet{wei-etal-2021-frequency} perform controlled experiments varying pretraining data to characterize the extent pretraining affects LMs' ability to learn to memorize and reason with facts as well as learn generalizable syntax rules.
In line with our results, both of these works find that frequency is a distinguishing factor in whether or not the model memorizes a particular fact or syntactic rule for a verb form.
\citet{sinha2021orderword} further demonstrate that shuffling word order during pretraining has minimal impact on an LMs' accuracy on downstream tasks, and, concurrent with this work, \citet{min2022rethinking} similarly find that shuffling labels in in-context learning demonstrations has a minimal impact on few-shot accuracy. These results further suggest that LMs' performance is largely driven by their ability to model high-order word co-occurrence statistics.
Although frequent terms are more likely to be memorized, data privacy researchers have also shown that LMs may memorize sensitive sequences occurring in training data (e.g., social security and credit card numbers), even if they are rare~\cite{carlini2019secret,song2019auditing}.

\paragraph{Memorization}
\citet{feldman2019does} provide a theoretical definition of memorization as the difference between the accuracy of a model on a training data point when that point is included vs. excluded from training.
In subsequent work, they develop an approach for approximating memorization using influence functions~\citet{feldman2020what}.
This framework is applied to study memorization in language models by \citet{zhang2021counterfactual}, who find that training examples that are memorized by the LM tend to have high influence of LM predictions on similar validation instances.
Their result may provide a plausible explanation that the frequency effects observed in this work are due to memorization.


\paragraph{Training Artifacts Challenge Evaluation}
Our results raise the issue that in-context learning probes may overestimate an LM's ability generalize from few examples when biases are present in the training data.
This is consistent with prior work that has exposed the similar effects of biases from: lexical cues in natural language inference datasets~\cite{gururangan2018annotation,poliak2018hypothesis,mccoy-etal-2019-right},
question-passage overlap and entity cues in reading comprehension datasets~\cite{chen2016thorough,sugawara2018easyrc,jia2017adversarial,lewis-etal-2021-question},
gender cues in coreference resolution datasets~\cite{rudinger2018gender},
popularity in named entity disambiguation~\cite{chen-etal-2021-evaluating}, similarity between training and test instances in information extraction and sentiment analysis datasets~\cite{elangovan-etal-2021-memorization}, 
and effects of how data is split~\cite{gorman-bedrick-2019-need,sogaard-etal-2021-need}.
Relatedly, data poisoning research studies how to adversarially introduce artifacts into training data to produce unwanted model behaviors~\cite{nelson2008exploting,chan2020poison,wallace2021concealed}.
A general statistical procedure to test for artifacts is presented in \citet{gardner2021competency}, who also theoretically show that large datasets are almost certain to contain artifacts under reasonable assumptions.
Techniques for mitigating biases in the presence of dataset artifacts are covered by \citet{romanov-etal-2019-whats} and \citet{karimi2020bias}.


\paragraph{Documenting Pretraining Data}
To better understand the risks of dataset artifacts, there has been a call to better document the characteristics and intended uses of datasets~\cite{gebru2021datasheets,bender2021dangers}.
However, due to the sheer size of LM pretraining datasets---which range from 100's of GBs to 10's of TBs---doing so can pose a substantial challenge.
Despite this, researchers have been able to estimate word frequencies, topics, and genres of documents~\cite{sharoff2020know}, as well as proportions of toxic text~\cite{gehman2020realtoxicityprompts} appearing in OpenWebText~\cite{Gokaslan2019OpenWeb}.
Similar efforts have been made to characterize the top-level domains, amount of hate speech, and censured text appearing in the C4 corpus~\cite{raffel2020exploring,dodge2021documenting,luccioni2021whats}.
Our work documents co-occurrence statistics of numbers and dates of documents appearing in the Pile dataset.

\paragraph{Numeracy and Temporal Reasoning in LMs}
Our work contributes a larger body of work dedicated to studying numeracy in word embeddings and language models~\cite{spithourakis-riedel-2018-numeracy,wallace2019numeracy}.
Recently, \citet{geva-etal-2020-injecting} and \citet{zhou-etal-2020-temporal} have proposed training schemes to help improve LMs' temporal and numerical reasoning capabilities.
\citet{patel2021nlp} also showed that NLP math solvers rely on simple heuristics to answer math questions. 
We expect that the \metric metric proposed in this work will be useful to better understand the impact of such schemes.

\section{Discussion and Future Work}


In this work, we consider how to conduct few-shot evaluations in light of the analysis with respect to the pretraining data.  Prior work has attempted to control for overlap between training or pretraining data and the test instances, but as we have seen, those methods are insufficient.  For example, \citet{brown2020language} measure the impact of removing instances from evaluation datasets that share 13-gram overlap with their pretraining data on GPT-3's accuracy, and also argue that the low occurrence of exact phrases such as ``NUM1 + NUM2 ='' and ``NUM1 plus NUM2'' in the pretraining data indicate that the model's strong performance on arithmetic tasks is likely due to factors other than memorization.  However, we have seen that model performance is impacted by much simpler statistical patterns, as small as unigram overlaps with the pretraining data.  

For these reasons, we strongly recommend that evaluation of reasoning capabilities should take the pretraining corpus into account, and any claims of reasoning can only be made after demonstrating robustness to the effect of pretraining.
LM benchmarks typically have no reference to the model's pretraining data. 
However, it is impossible to interpret few-shot performance on any benchmark without reference to information from the data that the LM was trained on.
One possible addition to future evaluation is the performance gap between high-frequency and low-frequency terms, perhaps including only terms that have been seen more than some threshold value.  
It is worth mentioning that, even a performance gap of $0$ is likely not sufficient to demonstrate a claim of reasoning capabilities---what exactly constitutes ``reasoning'' remains ill-defined---but it may be a necessary condition, and one that current models do not meet.

There are a few limitations to our study that open up avenues for future research.
We are not making a \emph{causal} claim here, and in general, there may be confounders that we have not eliminated in our study.
We recommend further research in investigating methods in causal inference and interventions during training to provide finer-grained analysis of the effect of pretraining.
Since our approach aggregates fairly simple patterns, the effect we observe might be stronger if a wider variety and complexity of patterns is considered in the pretraining corpus.
Similarly, our work was also limited to numerical reasoning tasks, and it would be worthwhile to consider how much other reasoning capability evaluations are impacted by the same effect, which could be measured using the {\metric} metric introduced here.
Defining appropriate instance terms for other reasoning tasks such as commonsense reasoning will be a challenging but important direction for future work. 
With the insights in this work and these recommendations, we hope to inspire further studies into the effect of pretraining on LM performance.

\section{Conclusion}
We show that in-context language model performance on numerical reasoning tasks can be impacted significantly by low-order co-occurrence statistics in the pretraining data, raising questions on the extent to which these models are actually \emph{reasoning} to solve these tasks.
These observations suggest the necessity for reconsidering and redefining the reasoning evaluation schemes for the large language models.
Further characterizing the impacting factors on the models reasoning capacities is also an important tasks for the community.
Most importantly, we suggest that the NLP community should not treat the pretraining data of the large language models as unknown black boxes. 
Overlooking the impact of the pretraining data can be misleading in evaluating the model reasoning skills. 

\section*{Acknowledgements}
We would like to thank 
Yanai Elazar,
Mukund Sundarajan,
Marco Tulio Ribeiro,
Eric Wallace,
Shivanshu Gupta,
Navid Salehnamadi,
Pouya Pezeshkpour,
and 
Dylan Slack 
for valuable discussions and feedback on this work.
This material is sponsored in part by the DARPA MCS program under Contract No. N660011924033 with the United States Office Of Naval Research, by an Amazon Research Award, and by awards IIS-2046873 and IIS-204098 from the National Science Foundation. 



\bibliography{refs}
\bibliographystyle{icml2022}



\end{document}